\documentclass{article}

\usepackage[final,nonatbib]{tackling_climate_workshop_style}

\usepackage[utf8]{inputenc} 
\usepackage[T1]{fontenc}    
\usepackage{hyperref}       
\usepackage{url}            
\usepackage{booktabs}       
\usepackage{amsfonts}       
\usepackage{nicefrac}       
\usepackage{microtype}      
\usepackage{mathtools}
\usepackage{xcolor}         
\usepackage{multirow}
\usepackage{adjustbox}

\newcommand{\bftab}{\fontseries{b}\selectfont}

\usepackage{amsmath,amsfonts,bm}

\def\eqref#1{equation~\ref{#1}}

\def\1{\bm{1}}

\def\rvx{{\mathbf{x}}}
\def\rvy{{\mathbf{y}}}
\def\rvz{{\mathbf{z}}}

\DeclareMathAlphabet{\mathsfit}{\encodingdefault}{\sfdefault}{m}{sl}
\SetMathAlphabet{\mathsfit}{bold}{\encodingdefault}{\sfdefault}{bx}{n}

\def\gX{{\mathcal{X}}}
\def\gY{{\mathcal{Y}}}
\def\gZ{{\mathcal{Z}}}

\usepackage[numbers]{natbib}
\title{Climate Variable Downscaling \\ 
	with Conditional Normalizing Flows}

\author{%
	Christina Winkler, Paula Harder, David Rolnick \\
	\texttt{ce.winkler@protonmail.com;\{paula.harder, david.rolnick\}@mila.quebec} \\
	\\
}

\begin{document}
	
	\maketitle
	\begin{abstract}
		Predictions of global climate models typically operate on coarse spatial scales due to the large computational costs of climate simulations. This has led to a considerable interest in methods for statistical downscaling, a similar process to super-resolution in the computer vision context, to provide more local and regional climate information. In this work, we apply conditional normalizing flows to the task of climate variable downscaling. We showcase its successful performance on an ERA5 water content dataset for different upsampling factors. Additionally, we show that the method allows us to assess the predictive uncertainty in terms of standard deviation from the fitted conditional distribution mean.
	\end{abstract}

	\section{Introduction}
	In climate modeling, simulations are typically run at coarse spatial resolution due to computational constraints. However, it is often of interest to obtain accurate predictions about the earth's climate not only on global but also on local scales, for example to guide local adaptation to precipitation or temperature trends. To fill this gap, statistical downscaling methods have increasingly been used to derive high-resolution information from low-resolution input. Early works have used Convolutional Neural Networks (CNNs) for climate variable downscaling \citep{Chen_2021, mital2022, harilal2021, Sha2020, geiss2020, liu2020, cheng2020}. However, these purely deterministic methods fail to capture the ill-determined nature of the problem -- for the same low-resolution image, there exist many possible fine-scale realizations. Capturing such stochasticity is important in order to improve the accuracy of local scale predictions. Therefore, Generative Adversarial Networks (GANs) have become a widely used method in super-resolution and climate variable downscaling \cite{Wang_2018_ECCV_Workshops, watson2020investigating,Climalign, Singh2019DownscalingNW, harder2022generating, watson2020investigating, Chaudhuri2020}. However, such methods lack latent-space encoders and are known to suffer from mode collapse. It is hard to assess whether they are overfitting or generalizing. In climate variable downscaling, we require estimating a density as close to the true high-resolution pixel distribution as possible, as high-frequency details are of main importance. In this field, recently exact likelihood methods such as diffusion models have been applied \cite{wan2023debias} for climate variable downscaling by leveraging theory from optimal transport. First, the data is debiased and then a diffusion model is used for upsampling. In \cite{Climalign}, the authors use normalizing flows for aligning the latent variables to a reference representation after performing statistical downscaling using a GAN architecture. Current state-of-the-art work \cite{yang2023fourier} uses Fourier neural operators to learn a mapping between high and low-resolution climate data for arbitrary resolution downscaling.
	
	In this work, we introduce the use of Conditional Normalizing Flows \cite{winkler2019learning} (CNFs) for stochastic climate variable downscaling. They are particularly desirable, since we can tractably compute likelihood values, their sampling procedure is efficient, and we are able to assess predictive uncertainty due to its probabilistic interpretation. Unlike other generative models where predictive uncertainty is often computed over an ensemble of different runs of weight initializations, or using techniques such as dropout, we are able to directly evaluate the predictive uncertainty of the CNF by computing the standard deviation from the fitted distribution mean.

	\paragraph{Contributions} Our main contributions can be summarized as follows:
	
	\begin{itemize}
		\item We show for the first time how to apply conditional normalizing flows to the task of climate variable downscaling.
		\item We verify that CNF makes it possible to evaluate predictive uncertainty, by computing uncertainty maps from the standard deviation of sampled outputs. 
	\end{itemize}
	
	\section{Background}\label{sec:background}
	\textbf{Normalizing flows} represent a function as the composition of simpler invertible functions $f(\mathbf{z}) = f^K \circ f^{K-1} \circ \cdots \circ f^1 (\mathbf{z})$ which yield the transformed random variables $\mathbf{z}^K \leftarrow ... \leftarrow \mathbf{z}^1 \leftarrow \mathbf{z}^0$ as intermediates after applying the transformations $f^1$ through $f^k$. The functions $f^k: \mathbb{R}^d \longmapsto \mathbb{R}^d$ are defined such that $f(\mathbf{z}^0) =\mathbf{y}$ with $\mathbf{y}\in \mathbb{R}^d$. All transformations $f^k$ are invertible and differentiable, making it possible to computing the Jacobian determinant. Then, by applying the \textit{Change of Variables Formula} we can model the density:
	
	\begin{equation}
	\begin{aligned}
	p_y(\rvy)& 
	&= p_\rvz(f(\rvy)) \begin{vmatrix}
	\det \frac{\partial f(\rvy)}{\partial \rvy}  \end{vmatrix}
	\end{aligned}\label{eq:change-of-variables}
	\end{equation}
	
	where $\rvy$ is our input data at training time mapping to latent variable $\rvz$. This allows us to formulate a model for the marginal likelihood $p_{\rvy}(\rvy)$ that can be computed tractably and optimized on the negative log-likelihood. We propose to learn conditional likelihoods using conditional normalizing flows \cite{winkler2019learning} for the task of super-resolution on climate data. Take as input the low-resolution image $\rvx \in \mathcal{X}$ and as target the high-resolution image $\rvy \in \mathcal{Y}$. We learn a distribution $p_{Y|X}(\rvy | \rvx)$ using a conditional prior $p_{Z|X}(\rvz | \rvx)$ and a mapping $f_\phi: \gY \times \gX \to \gZ$, which is bijective in $\gY$ and $\gZ$. The likelihood of this model is then defined as:
	\begin{equation}
	p_{Y|X}(\rvy | \rvx) = p_{Z|X}(\rvz | \rvx)  \left\lvert \frac{\partial \rvz}{\partial \rvy} \right\rvert = p_{Z|X}(f_{\phi}(\rvy , \rvx) | \rvx)  \left\lvert \frac{\partial f_{\phi}(\rvy , \rvx)}{\partial \rvy} \right\rvert. \label{eq:cnf}
	\end{equation}
	
	Notice that the difference between Equations \ref{eq:change-of-variables} and \ref{eq:cnf} is that all distributions are conditional, and the flow has a conditioning argument of $\rvx$. The generative process or in our case super-resolving an image from $\rvx$ to $\rvy$ is described by first sampling $\rvz \sim  p_{Z|X}(\rvz | \rvx)$ from a simple base density with its parameters conditioned on $\mathbf{x}$ (for us this is a diagonal Gaussian) and then passing it through a sequence of invertible mappings $f^{-1}_{\phi}(\rvz; \rvx)$ to obtain a predicted super-resolved image $\hat{\rvy}$. 
	
	\section{Experiments}
	
	
	
	\paragraph{ERA5 Hourly Water Content Dataset:} This reanalysis dataset measures Total Column Water (TWC) provided in $\frac{kg}{m^2}$. It describes the vertical integral of the total amount of atmospheric water content, that is, cloud water, water vapor, and cloud ice, but not precipitation. We use the same water content dataset as described in \cite{harder2022generating} who perform physically consistent downsampling to create the low-resolution image counterparts. The dataset includes 40,000 training samples, with 10,000 for validation and 10,000 for testing. Similar as before, for preprocessing, we transform the input data values $Z$ by $X=\frac{Z-\min{Z}}{\max{Z}-\min{Z}}$ such that they lie within range [0,1].

	\paragraph{Experimental setup:}
	For all experiments, we train the conditioned spatio-temporal flow with a learning rate of 2e-4 using a step-wise learning rate scheduler with a decay rate of 0.5 after every 200,000th parameter update step. We used the Adam optimizer \cite{kingma2014method} with exponential moving average and coefficients of running averages of gradients and its square are set to $\beta=(0.9,0.99)$. We train the model with an architecture of 3 scales and 2 flow steps per scale for 35 epochs.
	
	\subsection{Qualitative Evaluation}
\begin{figure}[tb]
		\centering
		\minipage{0.3\textwidth}
		\includegraphics[width=\linewidth]{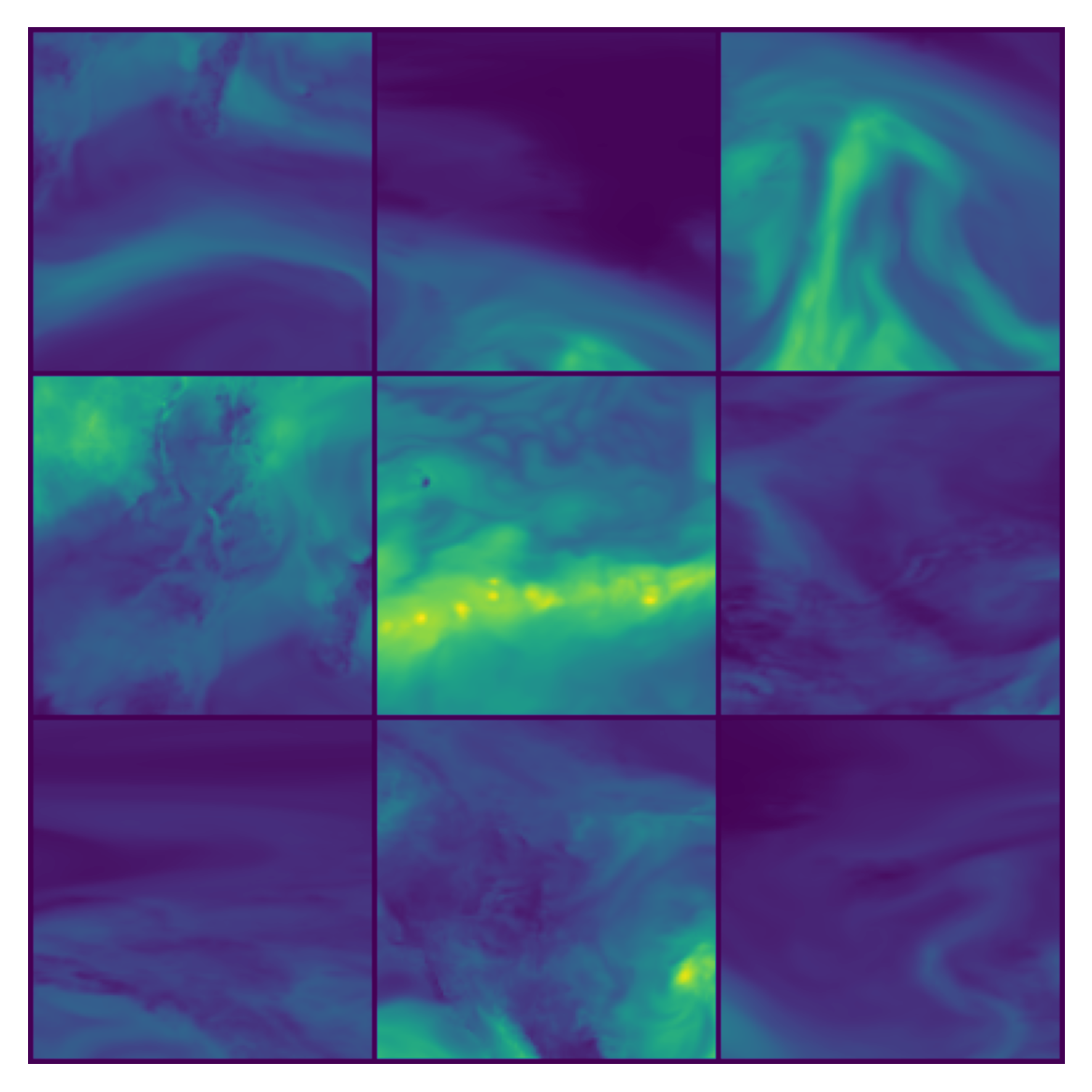}
		\centering
		(a) \textit{Ground truth}
		\endminipage 
		\minipage{0.3\textwidth}
		\includegraphics[width=\linewidth]{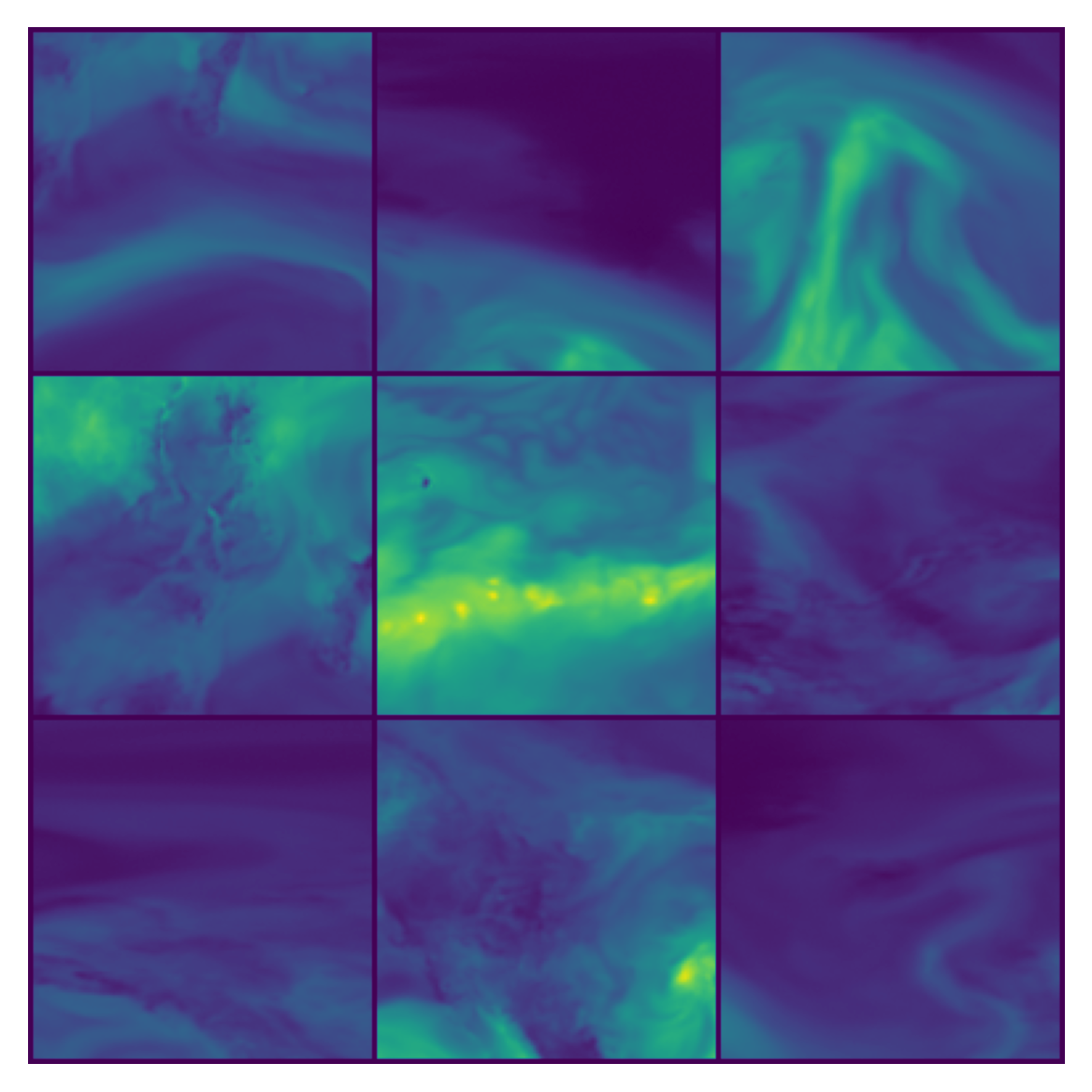}
		\centering
		(b) \textit{Super-resolved samples}
		\endminipage 
		\minipage{0.3\textwidth}%
		\includegraphics[width=\linewidth]{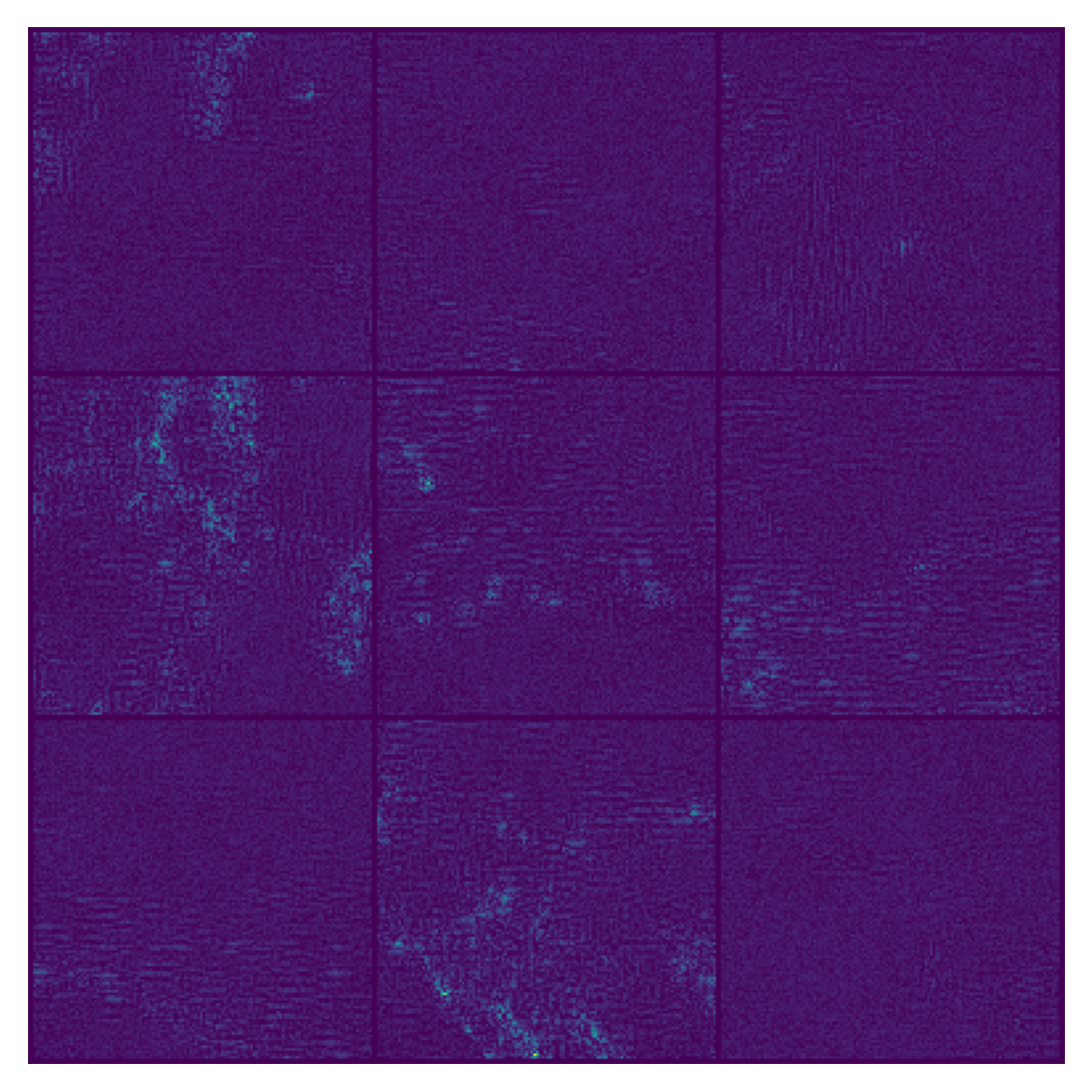}
		\centering
		(c) \textit{Absolute Error}
		\endminipage
		\caption{Super resolution results on the ERA5 water content TCW test data for 2 $\times$ upsampling. Samples are taken from the CNF $x_{hr} \sim p(x_{hr} | x_{lr})$ with $\tau=0.8$. \emph{Best viewed electronically.}}
		\label{fig:tcw-2x}
	\end{figure}

\begin{figure}[tb]
		\centering
		\minipage{0.3\textwidth}
		\includegraphics[width=\linewidth]{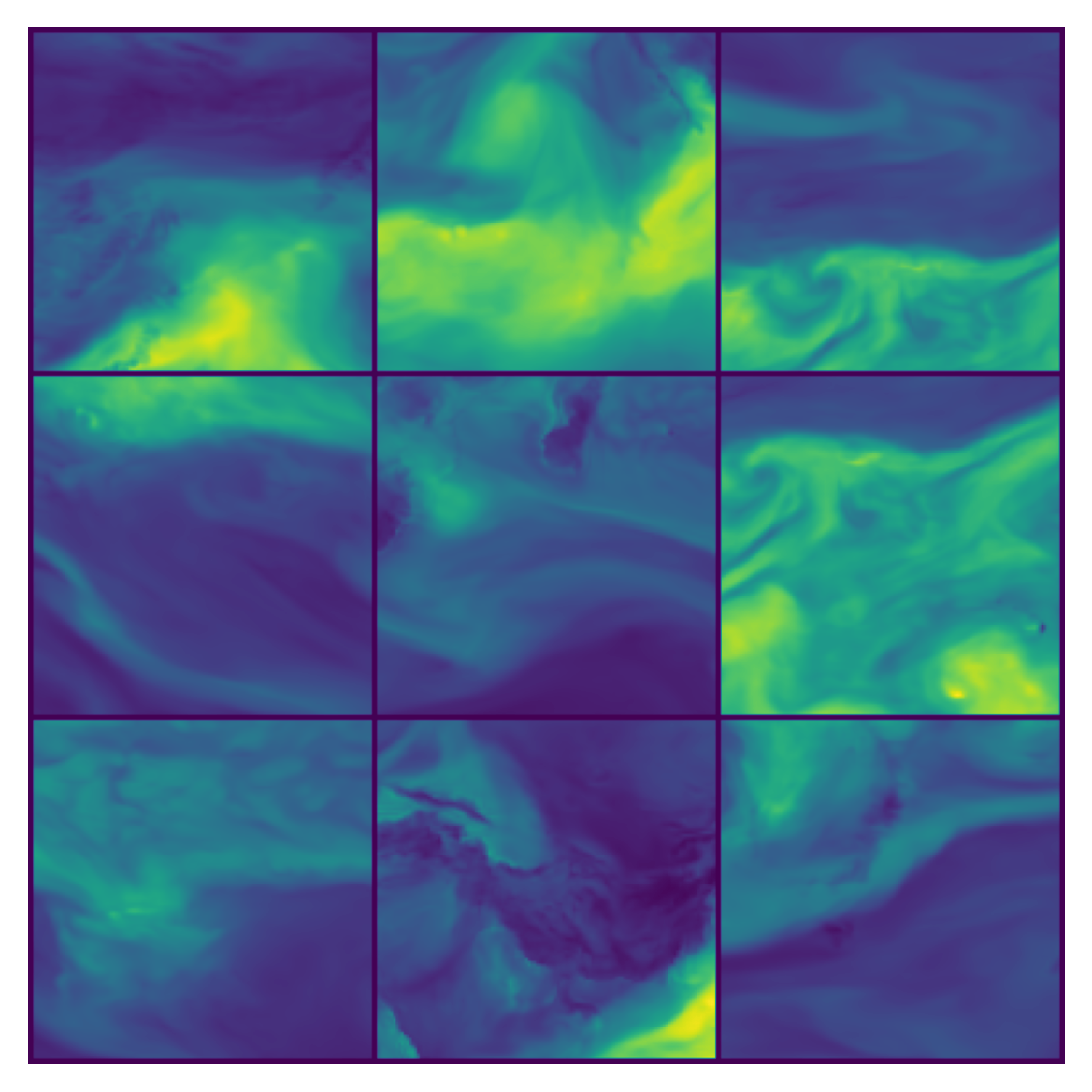}
		\centering
		(a) \textit{Ground truth}
		\endminipage 
		\minipage{0.3\textwidth}
		\includegraphics[width=\linewidth]{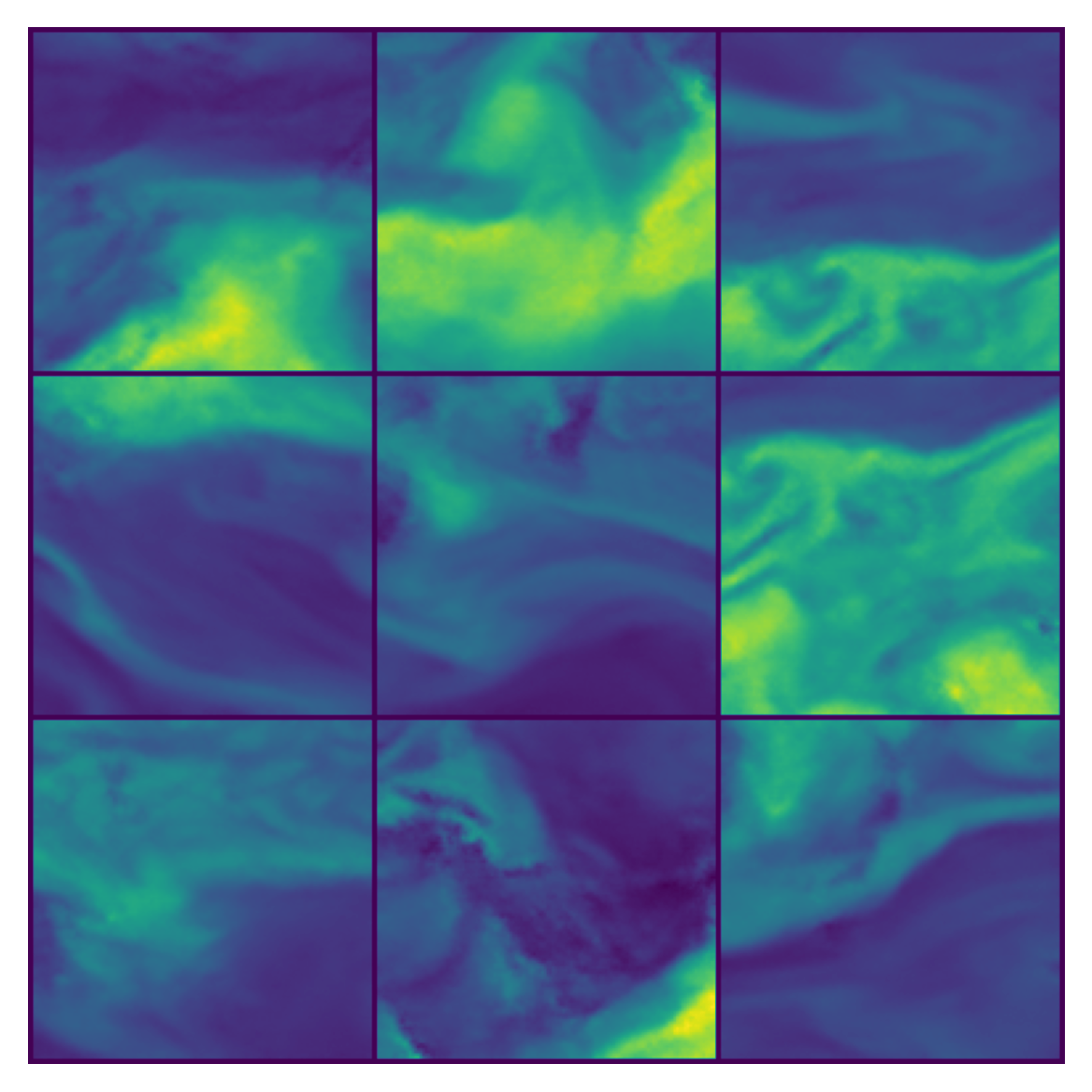}
		\centering
		(b) \textit{Super-resolved samples}
		\endminipage 
		\minipage{0.3\textwidth}%
		\includegraphics[width=\linewidth]{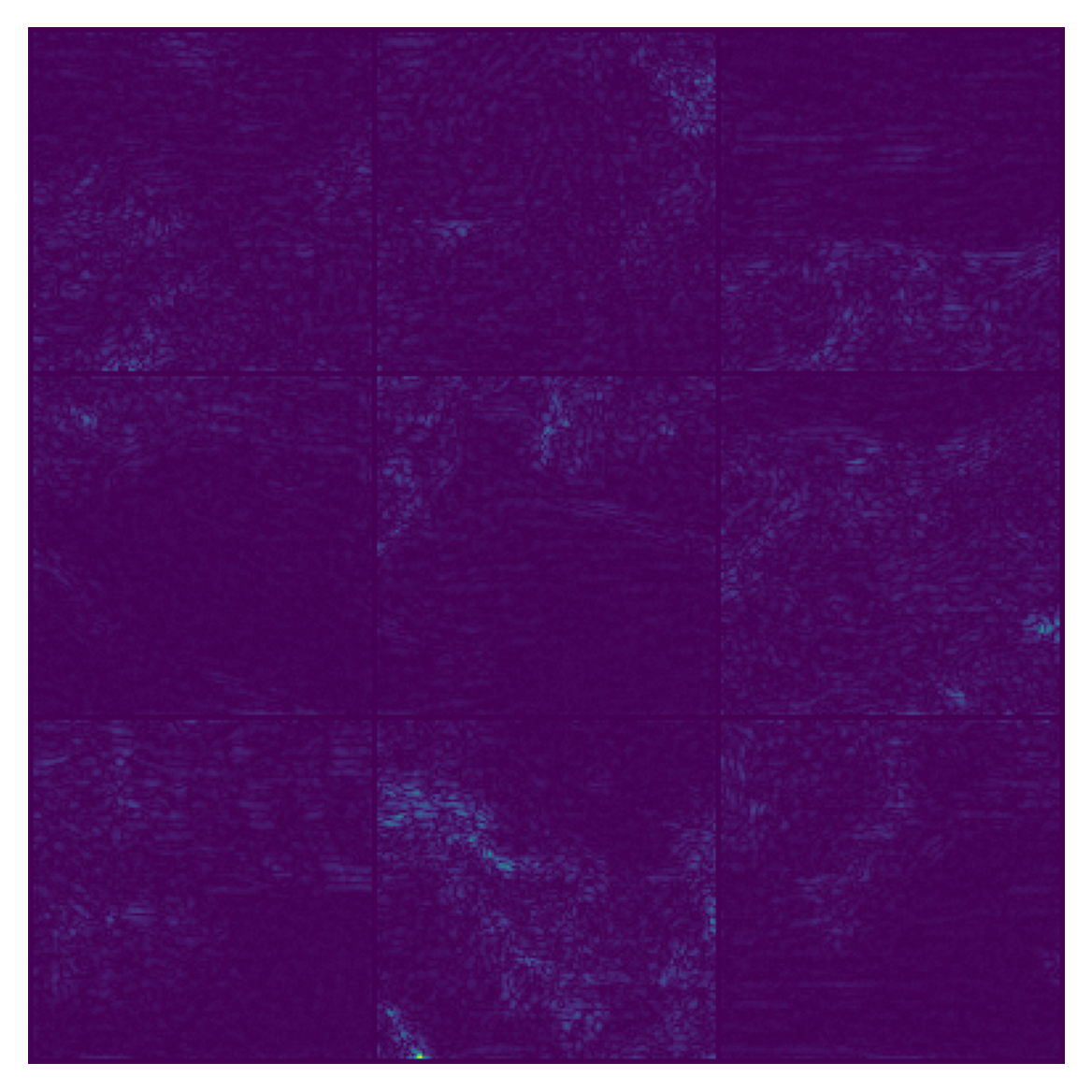}
		\centering
		(c) \textit{Absolute Error}
		\endminipage
		\caption{Super resolution results on the ERA5 TCW water content test data for 4 $\times$ upsampling. Samples are taken from the CNF $x_{hr} \sim p(x_{hr} | x_{lr})$ with $\tau=0.8$. \emph{Best viewed electronically.}}
		\label{fig:twc-4x}
	\end{figure}

	Figure \ref{fig:tcw-2x} and \ref{fig:twc-4x} display super-resolution results predicted by the conditional normalizing flow on the hourly water content and daily temperature datasets for upsampling factors of 2 and 4 respectively. The method is able to generalize over images in the test set, where each test sample conveys very different water content distributions. However, in regions with high intensity values, there is greater absolute error in predicting the correct pixel values than for regions with low intensity values. This may arise simply because the same percentage error results in a larger absolute error in such regions.
	
	\subsection{Quantitative Evaluation}
	Table \ref{table:metrics} shows the quantitative results of our method compared to a GAN architecture and bicubic interpolation. We added a perceptual Mean Squared Error loss between the predictions and ground truth image to improve sample quality. It can be seen that the generative approach outperforms the bicubic baseline. For the two times upsampling task, the super-resolution GAN outperforms the CNF on all metrics except the Continuous Ranked Probability Score (CRPS).
	
	
	
	
	
	
	
	
	
	\begin{table}[htb]
		\centering
		\caption{CNF evaluated on MAE, RMSE and CRPS on the held out ERA5 water content test set. We compare our method to bicubic interpolation and a GAN.}
		\label{table:metrics}
		\resizebox{\columnwidth}{!}{%
			\begin{tabular}[tb]{l|ccc|ccc}
				\toprule
				& \multicolumn{3}{c}{\bftab TCW 2$\times$ upsampling } &  \multicolumn{3}{c}{\bftab  TCW 4$\times$  upsampling} \\ 
				Model Type &
				MAE & RMSE & CRPS & MAE & RMSE & CRPS \\ \midrule
				Bicubic  
				\multirow{2}{*}{}
				& 6.96  $\pm$ 1.88 & 8.44 $\pm$ 2.09 & - &
				6.90 \ $\pm$ 3.04 & 8.37 $\pm$3.23  & - \\
				
				CNF 
				\multirow{2}{*}{} 
				& \bftab 5.22 $\pm$ 1.86 & \bftab 5.72  $\pm$ 1.98 & \bftab 0.0150 $\pm$ 0.0092 
				& \bftab 5.26  $\pm$ 1.86 & \bftab 5.80  $\pm$ 1.99 & \bftab 0.0174  $\pm$ 0.0118 \\
				
				CNF + Perc. Loss
				\multirow{2}{*}{} 
				& 5.29 $\pm$ 1.86 & 5.82 $\pm$ 1.99 &  0.0157 $\pm$ 0.0093
				& \bftab 5.25 $\pm$ 1.86 & \bftab 5.80 $\pm$ 1.98 & 0.0181 $\pm$ 0.0125  \\         
				
				
				CNF + Add. Constraint
				\multirow{2}{*}{} 
				& 5.26 $\pm$ 1.87 & 5.77 $\pm$ 2.0  & 0.0152 $\pm$ 0.0092
				& 5.54 $\pm$ 1.75  & 6.34 $\pm$ 1.76 &  0.0647 $\pm$ 0.0539\\    
				
				
				GAN \multirow{2}{*}{}
				& 5.27 $\pm$ 1.85  &  5.81 $\pm$ 1.97 & 0.0373 $\pm$ 0.0280 
				& 5.33  $\pm$ 1.84 & 5.90  $\pm$ 1.95 & 0.0454  $\pm$ 0.0393   \\
				
				\bottomrule
			\end{tabular}
		}
		
	\end{table}%
	
	
	
	
	
	
	
	
	\subsection{Sample Uncertainty}
	
	\begin{figure}
		\centering
		\includegraphics[width=\columnwidth]{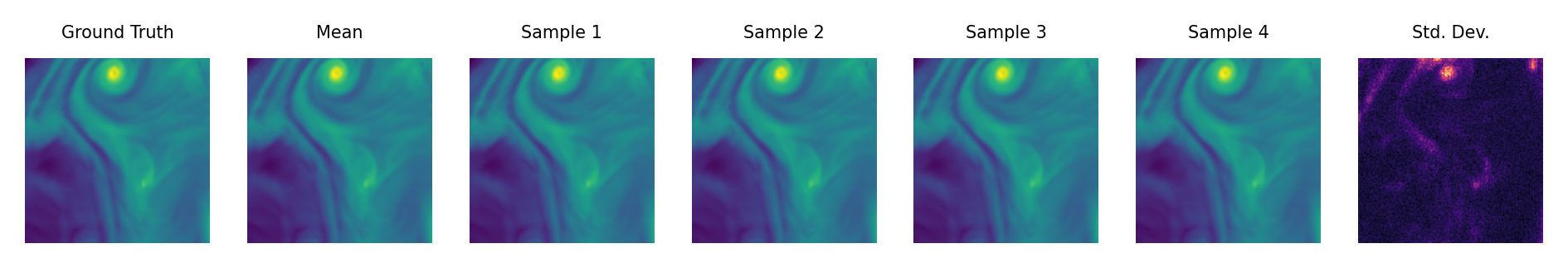}%
		\qquad
		\includegraphics[width=\columnwidth]{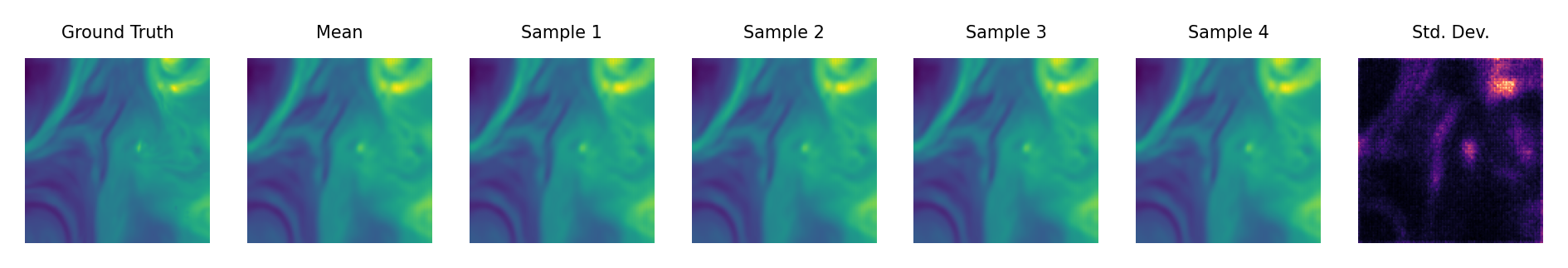}%
		\caption{The top row depicts the ground truth, conditional mean, different high-resolution realizations for one low-resolution image and computed standard deviation from the conditional mean for a 2$\times$ upsampling factor. The bottom row displays the same experiment for an upsampling factor of 4$\times$.}\label{fig:standard-dev}%
	\end{figure}
	
	One of the main advantages of normalizing flows is the ability to generate multiple samples for one initial condition. In our case, this would mean generating multiple high-resolution realizations for the same low-resolution image. Figure \ref{fig:standard-dev} visualizes the standard deviation computed across twenty samples from the model for one low-resolution image. For convenience, we plotted only four predicted samples to compare with. It can be seen that in areas of high variance and finer texture regions, the standard deviation is generally higher. In applications such as flood risk estimation, this may be highly advantageous, since we deliberately want to have a model which is able to capture anomalies in the water content distribution.
	
	
	

	\section{Conclusion} 
	In this work, we have shown the successful application of conditional normalizing flows to climate variables providing physically consistent results. The proposed method provides the advantage of density estimation and efficient sampling, and is able to model the stochasticity inherent in the relationships among fine and coarse spatial scales of climate variables. Additionally, we have shown that the method allows us to compute uncertainty maps in terms of standard deviation computed from the distribution mean. 
	
	
	\section{Acknowledgments}
	This research was supported by compute resources, software, and technical help generously provided by the Mila - Quebec AI Institute (\url{mila.quebec}). 
	
	\small
	\bibliography{bibliography}

\end{document}